\pdfoutput=1

\documentclass[11pt]{article}

\usepackage{ACL2023}

\usepackage{times}
\usepackage{latexsym}

\usepackage[T1]{fontenc}

\usepackage[utf8]{inputenc}

\usepackage{microtype}

\usepackage{inconsolata}
\usepackage{amsmath}
\usepackage{microtype}
\usepackage{array}
\usepackage{pifont}
\usepackage{tabularx}
\usepackage{adjustbox}
\usepackage{multirow}
\usepackage{enumitem}
\usepackage{xspace}
\usepackage{tcolorbox}
\usepackage{booktabs,subcaption,amsfonts,dcolumn}
\usepackage{hyperref}
\usepackage{url}
\usepackage{amsmath,amsthm,amsfonts,amssymb,bm,stmaryrd}
\usepackage{xcolor}
\usepackage{diagbox}

\newcommand\tf[1]{\textbf{#1}}
\newcommand\ttt[1]{\texttt{#1}}

\renewcommand{\paragraph}[1]{\vspace{0.2cm}\noindent\textbf{#1}}

\definecolor{midnightgreen}{rgb}{0.0, 0.29, 0.33}

\PassOptionsToPackage{hyphens}{url}\usepackage{hyperref}

\title{Augmentation-Adapted Retriever Improves Generalization of Language Models as Generic Plug-In}

\author{Zichun Yu$^1$ \quad Chenyan Xiong$^2$ \quad Shi Yu$^1$ \quad Zhiyuan Liu$^{13}$
\\ $^1$Dept. of Comp. Sci. \& Tech., Institute for AI, Tsinghua University, Beijing, China\\
$^2$Microsoft Research, Redmond, USA\\
$^3$Beijing National Research Center for Information Science and Technology, Beijing, China\\
\ttt{\{yuzc19, yus21\}@mails.tsinghua.edu.cn}; \ttt{chenyan.xiong@microsoft.com}\\
\ttt{liuzy@tsinghua.edu.cn}
}

\begin{document}
\maketitle

\begin{abstract}

Retrieval augmentation can aid language models (LMs) in knowledge-intensive tasks by supplying them with external information. 
Prior works on retrieval augmentation usually jointly fine-tune the retriever and the LM, making them closely coupled.
In this paper, we explore the scheme of generic retrieval plug-in: the retriever is to assist target LMs that may not be known beforehand or are unable to be fine-tuned together. To retrieve useful documents for unseen target LMs, we propose augmentation-adapted retriever (AAR), which learns LM's preferences obtained from a known source LM.
Experiments on the MMLU and PopQA datasets demonstrate that our AAR trained with a small source LM is able to significantly improve the zero-shot generalization of larger target LMs ranging from 250M Flan-T5 to 175B InstructGPT.
Further analysis indicates that the preferences of different LMs overlap, enabling AAR trained with a single source LM to serve as a generic plug-in for various target LMs. Our code is open-sourced at \href{https://github.com/OpenMatch/Augmentation-Adapted-Retriever}{https://github.com/OpenMatch/Augmentation-Adapted-Retriever}.

\end{abstract}
\section{Introduction}

Large language models (LMs) that possess billions of parameters are able to capture a significant amount of human knowledge, leading to consistent improvements on various downstream tasks~\cite{brown2020language,kaplan2020scaling,roberts2020much}.
However, the undeniable drawback of large LMs lies in their high computational cost, which negatively impacts their efficiency~\cite{strubell-etal-2019-energy,bender2021danger}.
Furthermore, the knowledge memorized from pre-training and the implicit reasoning process of LMs can be inaccurate and intractable sometimes, hindering their applications on knowledge-intensive tasks~\cite{guu2020realm,rag,mallen2023PopQA,wei2022chain}.

Instead of leveraging the knowledge and reasoning abilities embedded within the parameters of the LMs, \textit{retrieval augmentation}~\cite{guu2020realm,rag,pmlr-v162-borgeaud22a} enhances the LM with a retriever that can retrieve knowledge from an external corpus.
On the other hand, prior retrieval augmentation methods~\cite{izacard2020distilling,izacard_few-shot_2022} necessitate fine-tuning the backbone LM to adjust to the retriever and tackle specific downstream tasks. This kind of fine-tuning can be expensive when more and more unique demands emerge~\cite{maronikolakis-schutze-2021-multidomain}. More importantly, many top-tier LMs can only be accessed through black-box APIs~\cite{ouyang2022training,openai2023gpt4}. These APIs allow users to submit queries and receive responses but typically do not support fine-tuning.

\begin{figure}
    \centering
    \includegraphics[width=0.83\linewidth]{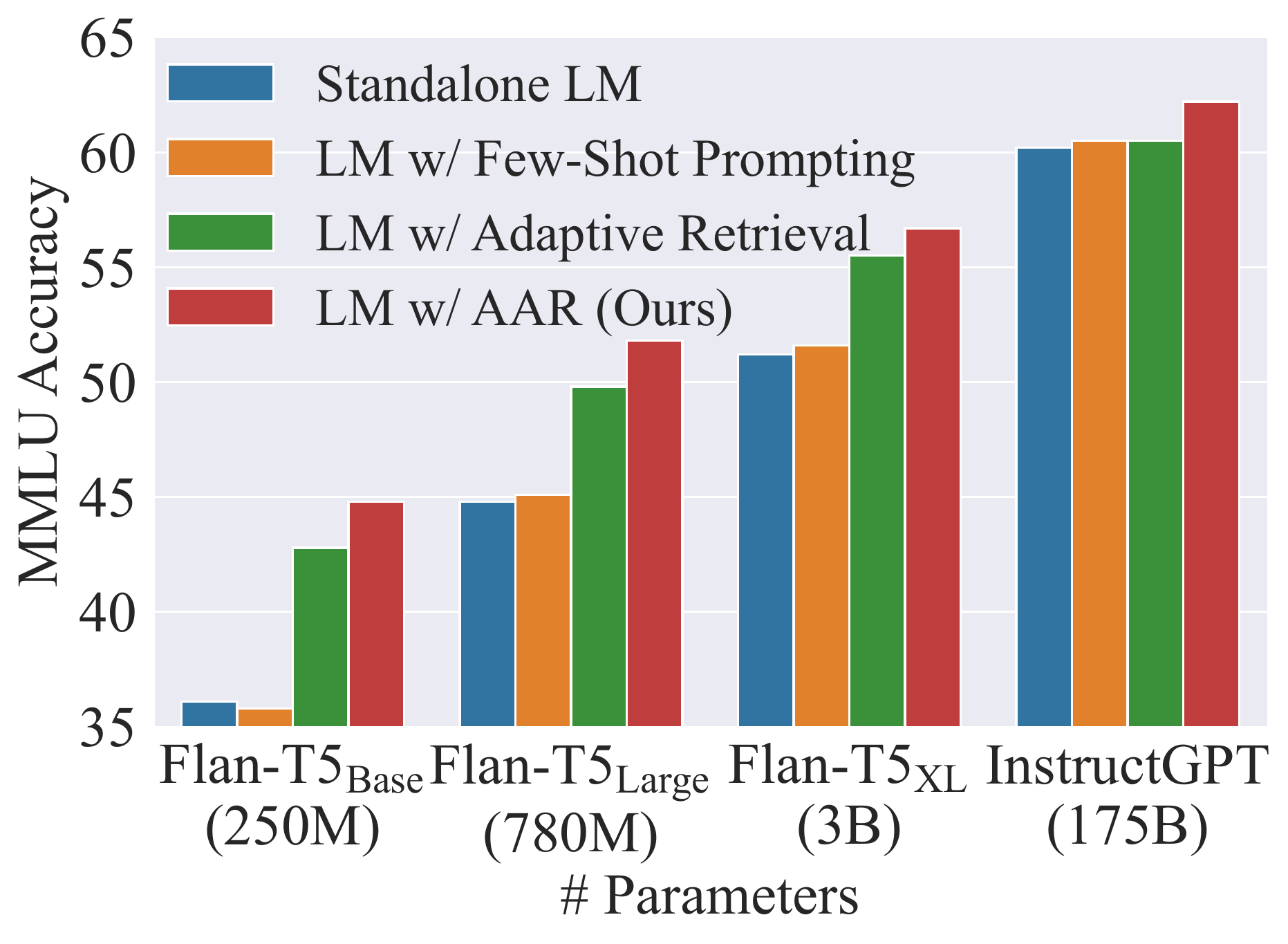}
    \caption{Performance of LM w/ AAR (Ours).}
    \label{fig-main}
\end{figure}

In this paper, we introduce \textbf{A}ugmentation-\textbf{A}dapted \textbf{R}etriever (AAR) to assist black-box LMs with downstream tasks as \textit{generic plug-in}.
To retrieve valuable documents for many unseen LMs, we propose to leverage a small \textit{source LM} to provide LM-preferred signals for retriever's training. The retriever after training (i.e., AAR) can be directly utilized to assist a large \textit{target LM} by plugging in the retrieved documents.

Specifically, we choose 
a small encoder-decoder LM as the source LM and utilize its fusion-in-decoder attention scores~\cite{izacard2020distilling} to annotate LM-preferred documents.
The LM-preferred documents are then combined with human-preferred documents to form the positive document set.
Negative documents are mined by the retriever itself using the ANCE~\cite{xiong2021approximate} technique.
After fine-tuning the retriever with LM's preferences, it can directly assist unseen target LMs in the zero-shot task generalization.


We evaluate AAR on a multi-task language understanding dataset MMLU~\cite{hendryckstest2021} and an entity-centric question answering dataset PopQA~\cite{mallen2023PopQA}. 
For the target LMs, we choose Flan-T5~\cite{flan-t5} series as our backbone for encoder-decoder LMs and InstructGPT~\cite{ouyang2022training} as our backbone for decoder-only LMs. Figure~\ref{fig-main} shows that assisted with a generic AAR, LMs of different sizes and architectures can consistently outperform the standalone LMs; the performance of smaller LMs can sometimes surpass the standalone counterparts of significantly larger sizes (e.g., Flan-T5$_{\text{Large}}$ w/ AAR outperforms standalone Flan-T5$_{\text{XL}}$ by 0.6\%).
AAR also demonstrates advantages over other augmentation approaches such as few-shot prompting and adaptive retrieval~\cite{mallen2023PopQA}. 

Further analysis reveals that the preferences obtained from different-sized source LMs are similar, and LMs with near capacities tend to yield closer preferred document sets. As a result, our AAR model trained from a small source LM can be considered as a generic plug-in to enhance the zero-shot generalization of a significantly larger target LM. We also discover that the documents preferred by LMs can provide assistance to the model from alternative perspectives, rather than relying solely on the full information favored by search users.
\section{Related Work}

\paragraph{Retrieval Augmentation.} 
Augmenting LMs with retrieved information from external memories has shown effective on diverse knowledge-intensive tasks~\cite{guu2020realm}.
Prior works explore novel ways to train the whole retriever-LM system in an end-to-end fashion, using retrieval-augmented sequence log-likelihood~\cite{rag,pmlr-v162-borgeaud22a}, fusion-in-decoder attention distillation~\cite{izacard2020distilling, izacard_few-shot_2022}, or knowledge graph~\cite{ju2022grape}. 
To decouple the retriever from LM, \citet{rubin-etal-2022-learning} train an independent prompt retriever for in-context learning, and \citet{Lin2022UnsupervisedCG} only fine-tune the LM via the retrieved data that is similar to few-shot unsupervised samples. 

Recent researches adopt zero-shot retrieval augmentation that does not fine-tune the LM on InstructGPT~\cite{ouyang2022training}. It can benefit entity-centric question answering~\cite{mallen2023PopQA}, chain-of-thought reasoning~\cite{he2023rethinking}, and multi-hop question answering~\cite{khattab2023dsp}. 
Parallel work~\cite{shi2023replug} uses LM likelihood to train the retriever for satisfying black-box LM's preferences, and they adopt GPT-3 Curie~\cite{brown2020language} to provide the supervision signals. 
In this work, we devise the retriever that can be used as a generic plug-in to assist a variety of unseen LMs.

\paragraph{Zero-shot Learning and Reasoning.} 
Large-scale unsupervised pre-trained LMs like GPT-3~\cite{brown2020language}, GPT-4~\cite{openai2023gpt4}, and PaLM~\cite{PaLM} are able to perform zero-shot learning on many downstream tasks with a task description provided at inference time.
Instruction-finetuned LMs~\cite{sanh2022multitask,flan-t5,ouyang2022training}, which are pre-trained on multiple supervised tasks using human instructions, also also exhibit robust zero-shot learning capabilities.
\citet{yu2022generate} propose a new scheme of zero-shot reasoning, which first prompts large LMs to generate relevant documents and then perform reading comprehension on the generated contents.
Recently, there has been a growing trend of utilizing plug-and-play knowledge injection to enhance the zero-shot performance of LMs, which is achieved through mapping network~\cite{zhang2023plug} or document encoding~\cite{xiao2023plug}.
Our work improves the zero-shot generalization of LMs by utilizing the retrieved information. We demonstrate that identifying LMs' preferences to train the retriever can in turn bring additional evidence texts for LMs.

\section{Method}
\label{sec:method}

In this section, we first introduce the preliminaries of the dense retrieval and the retrieval-augmented LM (§~\ref{sec:prelim}), then propose our augmentation-adapted retriever (§~\ref{sec:augmentation-adapted-retriever}).

\subsection{Preliminaries}\label{sec:prelim}

Retrieval-augmented LM~\cite{guu2020realm,rag} is a type of LM that leverages external information to improve its performance. It retrieves relevant documents from a corpus using a retriever, and then utilizes the documents to enhance its language generation capabilities.

The objective of the retriever is to find an augmentation document set $D^a$ from a corpus $C$ that helps the LM handle a given query $q$. Previous researches~\cite{karpukhin-etal-2020-dense,xiong2021approximate} concentrate primarily on the dense retrieval system that searches in the dense vector space since dense retrieval usually performs more accurately and efficiently than sparse one.

A dense retrieval model first represents $q$ and the document $d$ into an embedding space using a pre-trained encoder $g$,
\begin{equation}
    \setlength{\abovedisplayskip}{3pt}
    \setlength{\belowdisplayskip}{3pt}
    \bm{q} = g(q); \bm{d} = g(d), d \in C,
\end{equation}
and match their embeddings by dot product function $f$, which supports fast approximate nearest neighbor search (ANN)~\cite{andre2016cache, Johnson_Douze_Jegou_2021}. We then define $D^a$ that contains top-$N$ retrieved documents as:
\begin{equation}
    \setlength{\abovedisplayskip}{3pt}
    \setlength{\belowdisplayskip}{3pt}
    D^a = \{d_1^a \ldots d_N^a\} = \text{ANN}_{f(\bm{q}, \circ)}^N.
\end{equation}

For the LM backbones, the decoder-only and the encoder-decoder models are the two primary choices of the retrieval-augmented LMs~\cite{izacard2020leveraging,yu2022generate}.

Given a decoder-only LM like GPT-3~\cite{brown2020language}, the LM input can be a simple concatenation of the query and all the augmentation documents $\{d_1^a \ldots d_N^a\}$. Then, the LM will generate the answer based on the inputs auto-regressively.

For an encoder-decoder LM like T5~\cite{t5}, taking simple concatenation as the encoder input may still be effective. However, this method may not scale to a large volume of documents due to the quadratic self-attention computation associated with the number of documents. To aggregate multiple documents more efficiently, \citet{izacard2020leveraging} propose the fusion-in-decoder (FiD) mechanism, which soon becomes the mainstream in the development of encoder-decoder retrieval-augmented LMs. It first encodes each concatenation of the ($d_i^a$, $q$) pair separately and then lets the decoder attend to all parts:
\begin{equation}
    \setlength{\abovedisplayskip}{3pt}
    \setlength{\belowdisplayskip}{3pt}
    \text{FiD}(q) = \text{Dec}(\text{Enc}(d_1^a \oplus q)\ldots\text{Enc}(d_N^a \oplus q)).
\end{equation}

In this way, the encoder computes self-attention over one document at a time so that the computational cost can grow linearly with the number of documents. Furthermore, FiD cross-attention is found effective in estimating the relative importance of the augmentation documents from the LM's perspective~\cite{izacard2020distilling}. Therefore, soft FiD distillation~\cite{izacard2020distilling, izacard_few-shot_2022,shi2023replug}, which minimizes the KL-divergence between retrieval likelihood and LM likelihood, is often used to train the retriever and the LM end-to-end.

\begin{figure}
    \includegraphics[width=\linewidth]{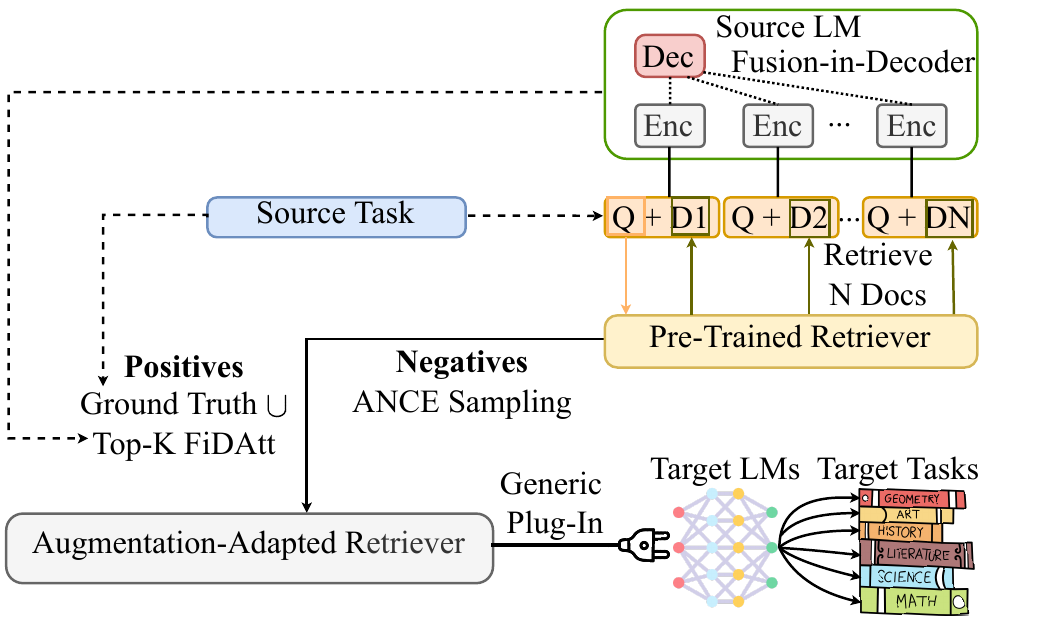}
    \centering
    \caption{Illustration of augmentation-adapted retriever.}
    \label{fig-method}
\end{figure}


\subsection{Augmentation-adapted Retriever}\label{sec:augmentation-adapted-retriever}


Due to the emerging real-world demands and the limitations of black-box APIs, fine-tuning retrieval-augmented LM for each possible downstream task can be infeasible. Hence, we introduce \textbf{A}ugmentation-\textbf{A}dapted \textbf{R}etriever (AAR) as a generic plug-in for black-box LMs. As illustrated in Figure~\ref{fig-method}, AAR can learn the preferences of LMs without the need for fine-tuning them.

Specifically, we utilize an encoder-decoder LM as source LM ($L_s$) to provide LM-preferred signals on a source task ($T_s$) for fine-tuning a pre-trained retriever. Then, we plug the fine-tuned retriever into unseen target LM ($L_t$) on a set of target tasks ($T_t$) non-intersecting with $T_s$.

Our training method starts from a source task $T_s$, where we aggregate the source LM $L_s$'s average FiD cross-attention (FiDAtt) scores $S_i^a$ corresponding to document $d_i^a$ from the first decoder token over all the layers, all the heads and all the input tokens $t$ of $d_i^a \oplus q$:
\begin{equation}\small
    S_i^a = \frac{1}{\text{ln} *\text{hn}*\text{tn}}\sum_{\text{layers}}\sum_{\text{heads}}\sum_{t \in d_i^a \oplus q}\text{FiDAtt}(\text{FiD}(q)).
\end{equation}
where ln, hn, tn are the numbers of the layers, the heads and the input tokens.

To make the training process more robust, we utilize the FiDAtt scores to annotate the LM-preferred positive documents in a discrete way:
\begin{equation}
    D^{a+} = D^{h+} \cup \text{Top-}K_{S_i^a,D^a},
\end{equation}
where $D^{h+}$ is the human-preferred positive document set (i.e., ground truth) on $T_s$. Top-$K_{S_i^a,D^a}$ means the documents with the top-k average FiDAtt scores $S_i^a$ in the retrieved document set $D^a$. 

Then, we sample hard negatives following ANCE~\cite{xiong2021approximate} and formulate the training loss $\mathcal{L}$ of the retriever as:

\vspace{-1.2em}
{\small
\begin{align}
    &D^{-} = \text{ANN}_{f(\bm{q}, \circ)}^M \backslash D^{a+},\\
    &\mathcal{L} = \sum_{q}\sum_{d^+ \in {D^{a+}}}\sum_{d^- \in {D^{-}}} l(f(\bm{q},\bm{d}^+),f(\bm{q},\bm{d}^-)),
\end{align}
}%
where $M$ is the hyperparameter of the negative sampling depth and $l$ is the standard cross entropy loss. After fine-tuning the retriever, we directly use it to augment unseen target LM $L_t$ on each task from target task set $T_t$.
\begin{table*}[t]
\centering
\resizebox{1.0\textwidth}{!}{%
\begin{tabular}{lll|ccccc|c}
\hline
\hline
    \multirow{2}{*}{\tf{Settings}} & \multirow{2}{*}{\tf{Methods}} & \multirow{2}{*}{\tf{\# Parameters}} & \multicolumn{5}{c|}{\tf{MMLU}} & {\tf{PopQA}} \\ & & & All & Hum. & Soc. Sci. & STEM & Other & All \\
\hline
\multicolumn{9}{l}{
\tf{Base Setting:} T5 Base Size} \\
\hline
Few-shot & Flan-T5$_{\text{Base}}$~\cite{flan-t5} & 250M & 35.8 & 39.6 & 39.8 & 26.3 & 41.2 & 8.0 \\
\hline
\multirow{4}{*}{Zero-shot} & Flan-T5$_{\text{Base}}$ & 250M & 36.1 & 40.4 & 39.8 & 27.0 & 40.6 & 8.8 \\
& Flan-T5$_{\text{Base}}$ w/ AR~\cite{mallen2023PopQA} & 250M & 42.8 & 43.5 & 44.0 & 35.8 & 50.0 & 29.4 \\
& Flan-T5$_{\text{Base}}$ w/ AAR$_{\text{Contriever}}$ (Ours) & 250M & 44.4 & \tf{44.7} & \tf{47.7} & 35.8 & 52.2 & 31.9 \\
& Flan-T5$_{\text{Base}}$ w/ AAR$_{\text{ANCE}}$ (Ours) & 250M & \tf{44.8} & 42.2 & 46.4 & \tf{39.0} & \tf{53.2} & \tf{37.7} \\
\hline
\multicolumn{9}{l}{
\tf{Large Setting:} T5 Large Size} \\
\hline
\multirow{2}{*}{Few-shot} & Atlas$_\text{Large}$ FT~\cite{izacard_few-shot_2022} & 770M & 38.9 & 37.3 & 41.7 & 32.3 & 44.9 & n.a. \\ 
& Flan-T5$_{\text{Large}}$ & 780M & 45.1 & 47.7 & 53.5 & 34.4 & 49.2 & 9.3 \\
\hline
\multirow{4}{*}{Zero-shot} & Flan-T5$_{\text{Large}}$ & 780M & 44.8 & 46.3 & 51.4 & 34.8 & 50.6 & 7.2 \\
& Flan-T5$_{\text{Large}}$ w/ AR & 780M & 49.8 & 50.0 & 55.6 & 38.4 & 59.5 & 29.6 \\
& Flan-T5$_{\text{Large}}$ w/ AAR$_{\text{Contriever}}$ (Ours) & 780M & \tf{51.8} & \tf{50.8} & \tf{59.7} & \tf{39.4} & \tf{61.8} & 33.4 \\
& Flan-T5$_{\text{Large}}$ w/ AAR$_{\text{ANCE}}$ (Ours) & 780M & 50.4 & 48.0 & 58.1 & 39.3 & 60.2 & \tf{39.3} \\
\hline
\multicolumn{9}{l}{
\tf{XL Setting:} T5 XL Size} \\
\hline
\multirow{2}{*}{Few-shot} & Atlas$_\text{XL}$ FT & 3B & 42.3 & 40.0 & 46.8 & 35.0 & 48.1 & n.a. \\ 
& Flan-T5$_{\text{XL}}$ & 3B & 51.6 & 55.0 & 61.1 & 36.8 & 59.5 & 11.1 \\
\hline
\multirow{4}{*}{Zero-shot} & Flan-T5$_{\text{XL}}$ & 3B & 51.2 & 55.5 & 57.4 & 38.1 & 58.7 & 11.3 \\
& Flan-T5$_{\text{XL}}$ w/ AR & 3B & 55.5 & 56.7 & 64.5 & 43.0 & 62.6 & 33.7 \\
& Flan-T5$_{\text{XL}}$ w/ AAR$_{\text{Contriever}}$ (Ours) & 3B & \tf{56.7} & 57.7 & \tf{65.4} & \tf{43.6} & \tf{65.1} & 31.5 \\
& Flan-T5$_{\text{XL}}$ w/ AAR$_{\text{ANCE}}$ (Ours) & 3B & 56.2 & \tf{59.4} & 64.8 & 41.5 & 64.9 & \tf{38.0} \\
\hline
\multicolumn{9}{l}{
\tf{Giant Setting:} Over 70B Size} \\
\hline
\multirow{3}{*}{Few-shot} & Chinchilla~\cite{Chinchilla} & 70B & 67.5 & 63.6 & 79.3 & 55.0 & 73.9 & n.a. \\
& OPT-IML-Max~\cite{OPT-IML} & 175B & 47.1 & n.a. & n.a. & n.a. & n.a. & n.a. \\	
& InstructGPT~\cite{ouyang2022training} & 175B & 60.5 & 62.0 & 71.8 & 44.3 & 70.1 & 35.2 \\
\hline
\multirow{6}{*}{Zero-shot} & GAL~\cite{GALACTICA} & 120B & 52.6 & n.a. & n.a. & n.a. & n.a. & n.a. \\
& OPT-IML-Max & 175B & 49.1 & n.a. & n.a. & n.a. & n.a. & n.a. \\
& InstructGPT & 175B & 60.2 & \tf{65.7} & 68.0 & 46.1 & 66.5 & 34.7 \\
& InstructGPT w/ AR & 175B & 60.5 & 62.2 & 71.3 & 44.7 & 69.7 & 43.3 \\
& InstructGPT w/ AAR$_{\text{Contriever}}$ (Ours) & 175B & 61.5 & 64.5 & \tf{73.1} & 45.0 & 69.9 & 43.9 \\
& InstructGPT w/ AAR$_{\text{ANCE}}$ (Ours) & 175B & \tf{62.2} & 62.0 & 72.0 & \tf{49.2} & \tf{70.7} & \tf{52.0} \\
\hline
\hline
\end{tabular}
}

\caption{
Our main results on MMLU and PopQA dataset. We group the methods mainly by the parameters. Our $L_s$ is Flan-T5$_{\text{Base}}$. AAR$_\text{Contriever}$: AAR initialized from Contriever; AAR$_\text{ANCE}$: AAR initialized from ANCE; FT: fine-tuning; AR: adaptive retrieval. Unspecified methods represent direct prompting. The score marked as \tf{bold} means the best performance among the models in the zero-shot setting. 
}
\label{tab:main}
\end{table*}

\section{Experimental Methodologies}

In this section, we discuss our main experimental setup. More details can be found in Appendix~\ref{sec:setting}.

\subsection{Target Tasks}

Following prior works~\cite{flan-t5,mallen2023PopQA}, we choose MMLU~\cite{hendryckstest2021} and PopQA~\cite{mallen2023PopQA} as target tasks $T_t$.

\paragraph{MMLU} is a multitask language understanding dataset, which includes 57 multi-choice question answering subtasks. These subtasks can be generally classified into four categories: humanities, social sciences, STEM, and other. We average the accuracy of the subtasks in each category to obtain the final score. We report the accuracy of the evaluation set in our main experiments.

\paragraph{PopQA} is an entity-centric question answering dataset that mainly concentrates on long-tail questions. We report the accuracy of the test set in our main experiments.

\subsection{Our Method}

\paragraph{Retrievers.} We adopt two widely used retrievers to initialize AAR: ANCE initialized from T5$_\text{Base}$~\cite{t5,ge2023augmenting} and Contriever~\cite{izacard2021contriever} initialized from BERT$_\text{Base}$~\cite{devlin-etal-2019-bert}. Both of them have been fine-tuned on MS MARCO~\cite{MSMARCO} previously. For the retrieval corpus, we choose the MS MARCO~\cite{MSMARCO} for MMLU and the KILT-Wikipedia~\cite{fb_kilt} for PopQA.

\paragraph{Language Models.} We adopt Flan-T5~\cite{flan-t5} series as our backbone for encoder-decoder LMs and InstructGPT\footnote{We use the $\text{GPT-3}_{\text{text-davinci-002}}$ December 2022 version.}~\cite{ouyang2022training} as our backbone for decoder-only LMs. These models have been multi-task instruction-finetuned and are widely utilized for assessing zero-shot generalization~\cite{zhou2023comprehensive}.

\paragraph{Implementation Details.} We utilize the MS MARCO~\cite{MSMARCO} as our source task $T_s$ since it is the common choice to train the retriever~\cite{xin-etal-2022-zero}. This dataset consists of high-quality questions that require real-world knowledge to answer, which aligns strongly with our target tasks $T_t$ and possesses no overlap with them.
Considering the implementation efficiency, we take the Flan-T5$_{\text{Base}}$ as the source LM $L_s$ and treat the larger model as the target LM $L_t$. We directly set the total document number $N=10$, LM-preferred document number $K=2$, and the negative mining depth $M=100$ in the augmentation-adapted training. We run all experiments on a single A100 GPU (40G).

\subsection{Baselines}

\paragraph{Zero-shot Setting.} We compare our method with the state-of-the-art zero-shot baselines. Standalone LMs, including Flan-T5~\cite{flan-t5}, InstructGPT~\cite{ouyang2022training}, GAL~\cite{GALACTICA} and OPT-IML-Max~\cite{OPT-IML}, are prompted by a natural language instruction that describes the desired task and question. Adaptive retrieval~\cite{mallen2023PopQA} selectively utilizes non-parametric memory (retrieval augmentation) and parametric memory (the knowledge obtained from pre-training) based on questions' popularity. In our main experiment, we select the optimal combination in their paper, which consists of Contriever as the non-parametric memory and GenRead~\cite{yu2022generate} as the parametric memory.

\paragraph{Few-shot Setting.} We also include the results of previous few-shot models for reference. Flan-T5, InstructGPT, Chinchilla~\cite{Chinchilla} and OPT-IML-Max adopt few-shot demonstrations, which provide the LMs with a limited number of task examples. This enables the models to generalize from these examples and generate accurate responses~\cite{gao2021making}. Atlas~\cite{izacard_few-shot_2022} is a state-of-the-art retrieval-augmented LM, which jointly pre-trains the retriever with the LM using unsupervised data and fine-tunes the retriever via the attention distillation on few-shot data.

\begin{figure}[t]
	\centering
    \includegraphics[width=1.0\linewidth]{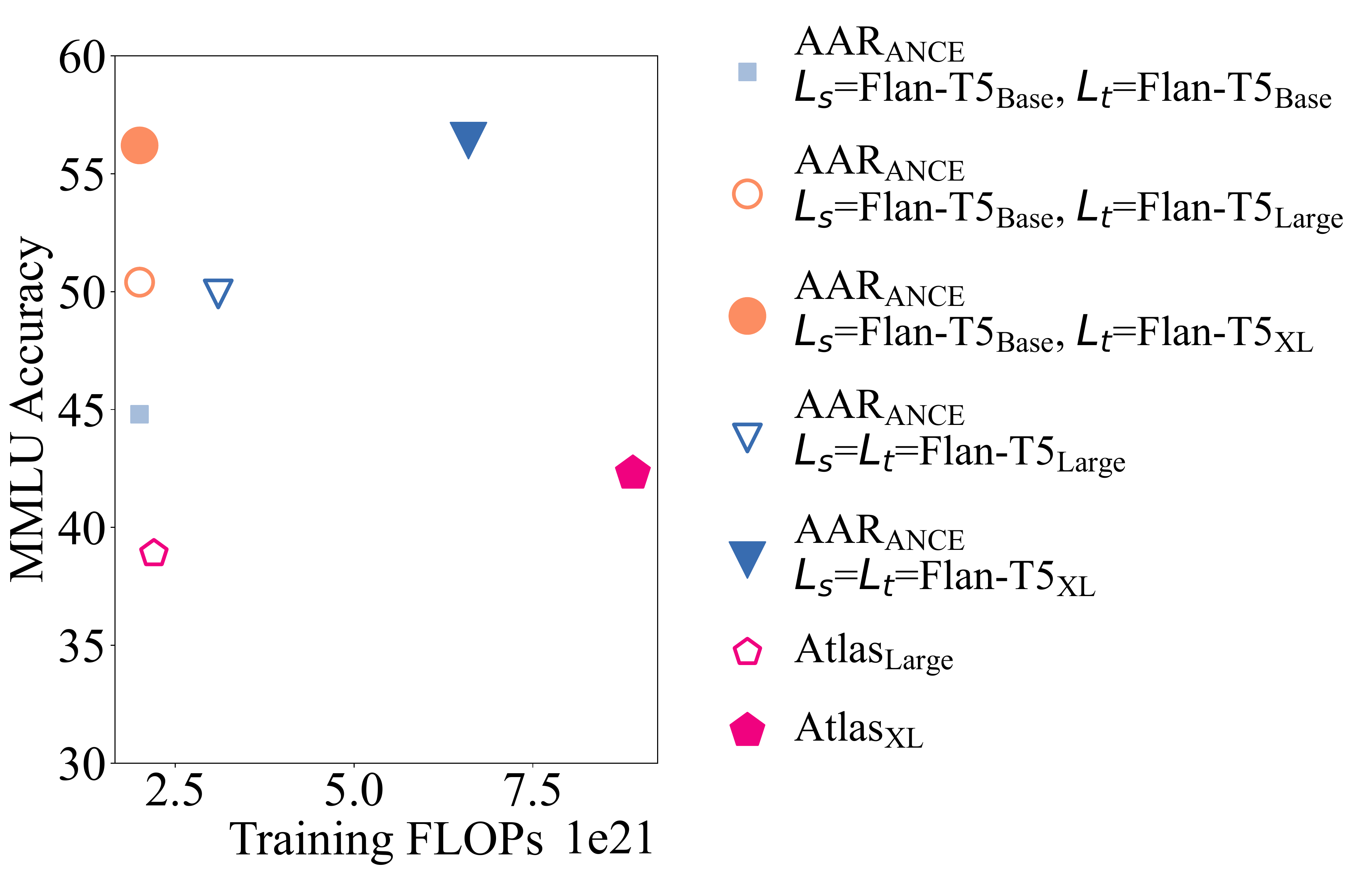}
    \caption{Training FLOPs of retrieval augmentation methods.}
    \label{fig-efficiency}
    \vspace{-1.5em}
\end{figure}
\section{Evaluation Results}
In this section, we discuss our main results on MMLU and PopQA datasets (§~\ref{sec:main-results}) and conduct comprehensive studies about how (§~\ref{sec:aat}, §~\ref{sec:LM-preferred documents}, §~\ref{sec:multi-task training}) and when (§~\ref{sec:corpus}, §~\ref{sec:scenarios}) AAR helps.

\subsection{Overall Performance}
\label{sec:main-results}

Table~\ref{tab:main} demonstrates that, with the assistance of a generic AAR, target LMs of different sizes and architectures can significantly outperform their standalone baselines in the zero-shot setting. Notably, AAR even improves powerful InstructGPT by 2\% on MMLU and by nearly 20\% on PopQA. We hypothesize that the PopQA dataset mainly comprises long-tail questions and thus necessitates more augmentation information to attain high accuracy. AAR outperforms other augmentation methods like few-shot prompting and adaptive retrieval, as they may not offer as extensive evidence text as AAR does.

Meanwhile, AAR is a highly efficient augmentation approach since it only relies on a small source LM Flan-T5$_{\text{Base}}$ (250M) to provide training signals and can generalize well to target LMs of larger capacities. Figure~\ref{fig-efficiency} illustrates that solely setting the source LM as the target LM (represented by the inverted triangles) does not significantly enhance the MMLU accuracy. However, it may triple the training budget required. Only using a small source LM is able to outperform the powerful Atlas by large margins with fewer training FLOPs.

\begin{figure}[t]
	\centering
	\begin{subfigure}{0.208\textwidth}
	    \centering
        \includegraphics[width=1.0\linewidth]{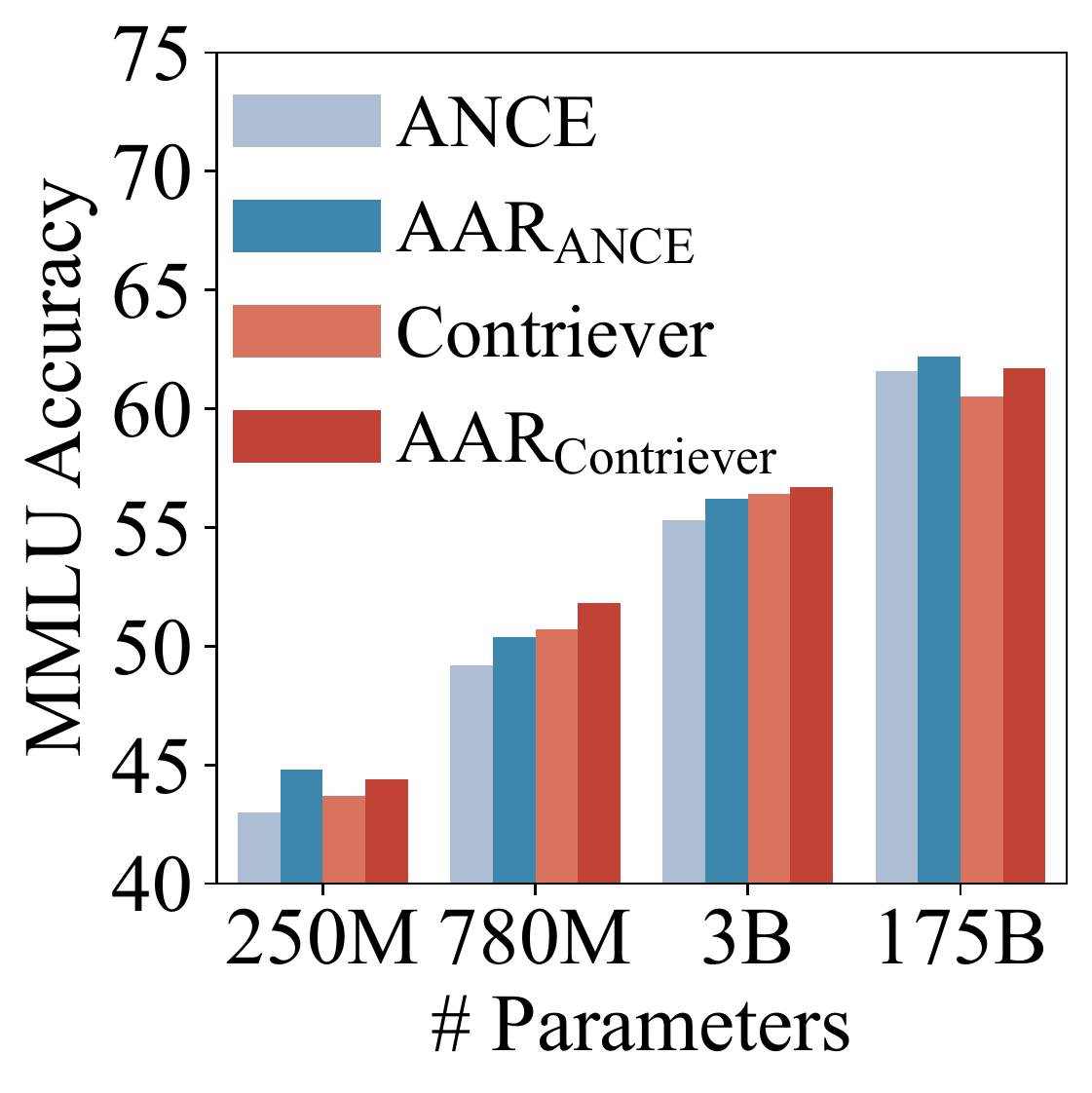}
        \caption{Pre-trained Retrievers.}
        \label{fig-aar-improvement}
	\end{subfigure}%
    \hspace{1pt}
	\centering
	\begin{subfigure}{0.208\textwidth}
		\centering
        \includegraphics[width=1.0\linewidth]{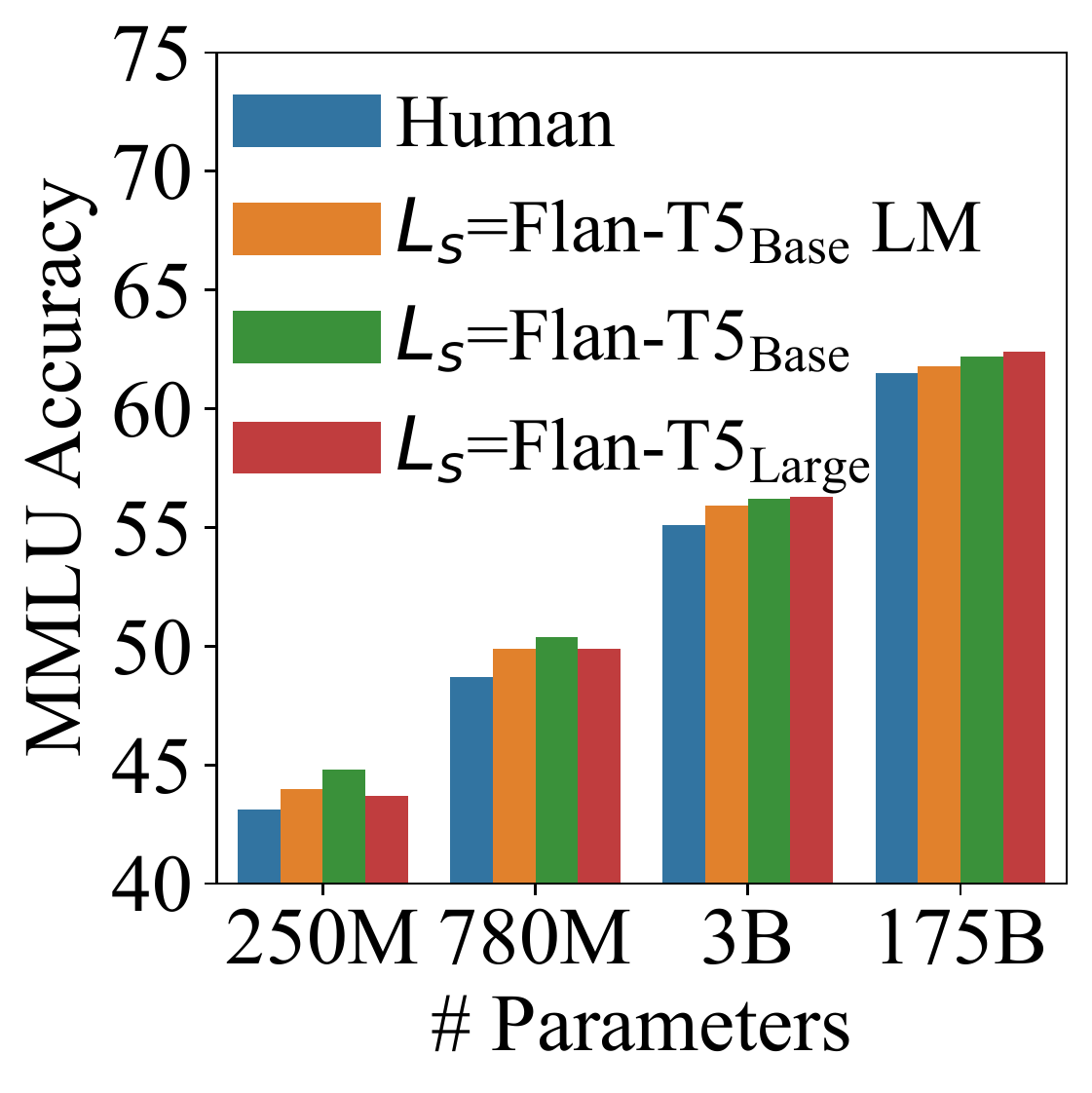}
        \caption{Positive docs selection.}
        \label{fig-positive-selection}
	\end{subfigure}%
    \caption{AAR's performance when (a) using different pre-trained retrievers and (b) trained with different positive documents, using Flan-T5$_{\text{Base}}$ (250M), Flan-T5$_{\text{Large}}$ (780M), Flan-T5$_{\text{XL}}$ (3B), InstructGPT (175B) as $L_t$. The retriever in (b) is initialized from ANCE.}
\end{figure}

\subsection{Ablation Study}
\label{sec:aat}

In this experiment, we conduct the ablation study of augmentation-adapted training and analyze model behaviors during the training process.

Figure~\ref{fig-aar-improvement} illustrates that augmentation-adapted training can bring additional improvements compared to the pre-trained retrievers. In general, ANCE benefits more from augmentation-adapted training than Contriever. This may be due to the fact that Contriever has been already intensively pre-trained on massive data augmentations as well as MS MARCO whereas ANCE is trained only on MS MARCO. We provide exact numbers in Table~\ref{tab:ft} and PopQA results in Figure~\ref{fig-popqa-improvement}, which yield similar observations as MMLU.

In Figure~\ref{fig-positive-selection}, we compare retrievers trained with different positive documents, including human-preferred documents annotated by search users (the blue bar), LM-preferred documents obtained by the source LM (the orange bar), and their combinations (the green bar and the red bar). Since the retriever has been pre-trained on user-annotated MS MARCO, simply using human-preferred documents to train it may be meaningless and therefore performs the worst among all approaches. Only using LM-preferred documents demonstrates notable gains over only using human-preferred documents, and merging both human-preferred and LM-preferred documents (our main setup) further enhances the retriever's performance.
Finally, using Flan-T5$_\text{Base}$ as source LM yields better results compared to using Flan-T5$_\text{Large}$ when the target LMs are relatively small. However, as the target LM's size increases, both approaches achieve comparable performance. Hence, our choice to utilize a small source LM in the augmentation-adapted training is reasonable and effective.

\begin{figure}[t]
	\centering
	\begin{subfigure}{0.246\textwidth}
	    \centering
        \includegraphics[width=1.0\linewidth]{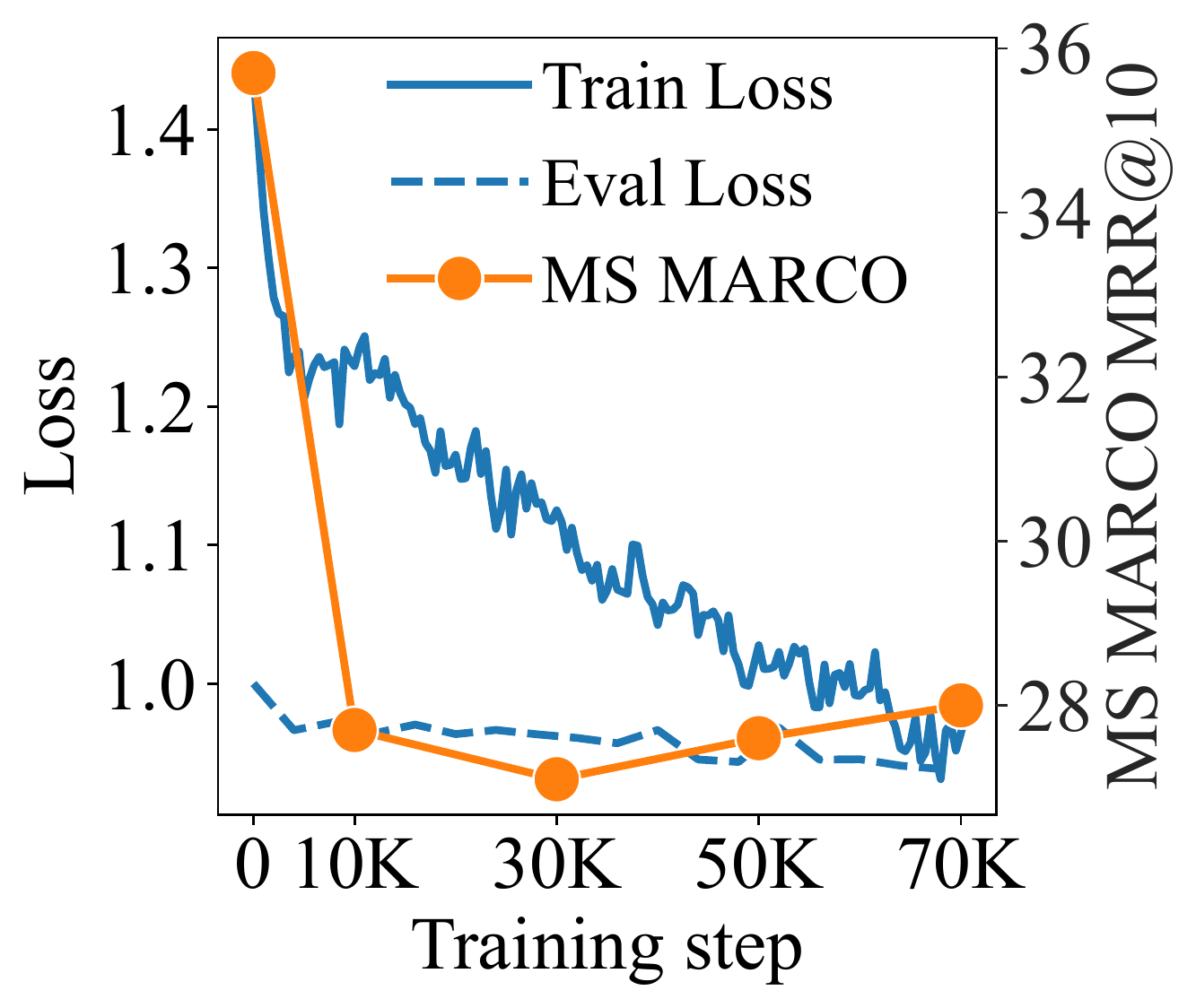}
        \caption{Retriever's performance.}
        \label{fig-retriever-training}
	\end{subfigure}%
	\centering
	\begin{subfigure}{0.254\textwidth}
		\centering
        \includegraphics[width=1.0\linewidth]{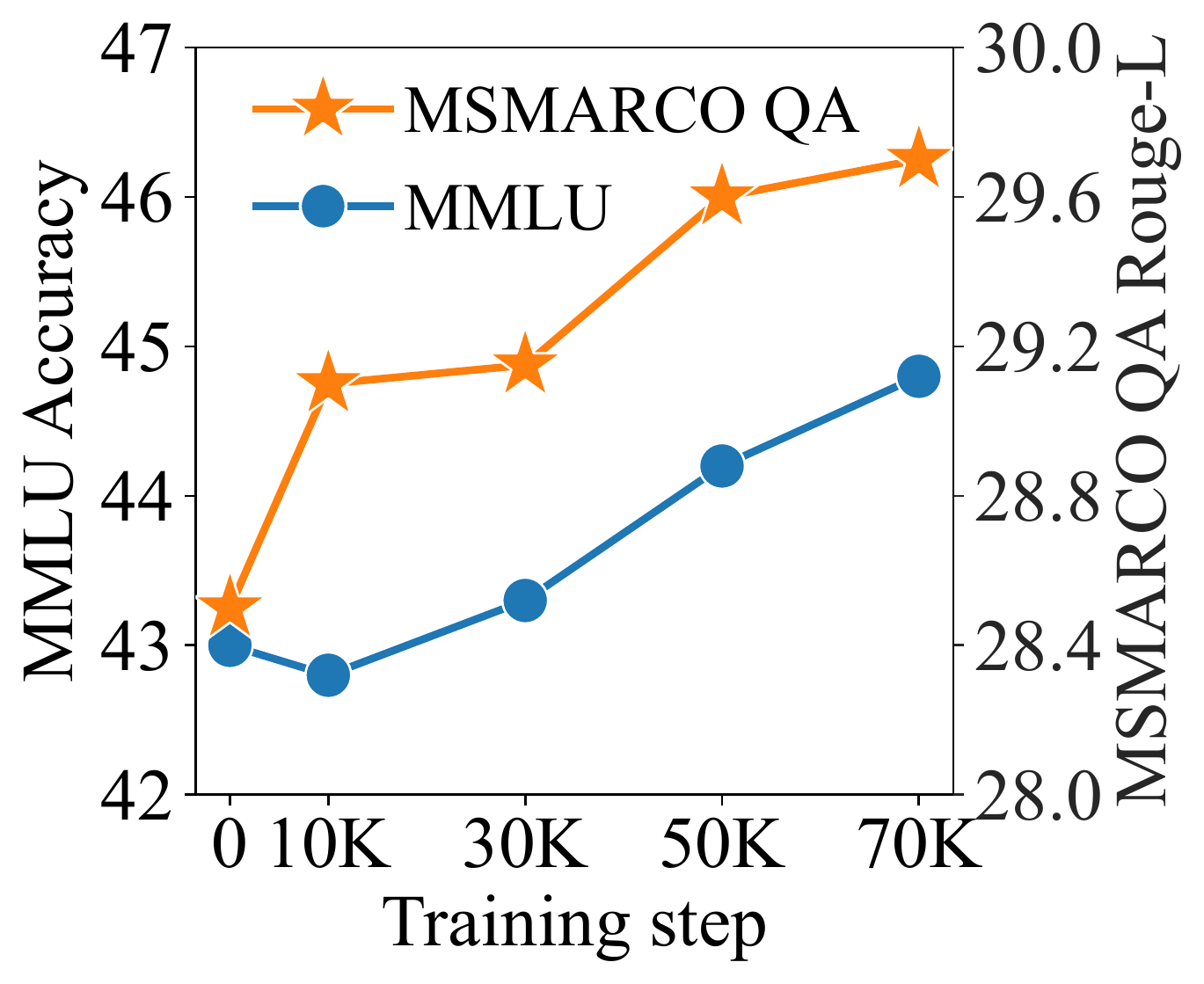}
        \caption{$L_t$'s performance.}
        \label{fig-LM-training}
	\end{subfigure}
	\caption{AAR's training process. (a) exhibits the retriever's (ANCE) performance on MS MARCO. (b) presents the $L_t$'s (Flan-T5$_{\text{Base}}$) performance on MSMARCO QA and MMLU.}
\end{figure}

Figure~\ref{fig-retriever-training} and Figure~\ref{fig-LM-training} plot the retriever's and LM's performance during augmentation-adapted training, respectively. At the beginning of the training, the retriever's MRR@10 on the MS MARCO drops dramatically, indicating a large distribution gap between human-preferred and LM-preferred documents. As the retriever's train and dev loss continually decline, the retrieval-augmented LM gradually performs better on MSMARCO QA and eventually, on MMLU. This result implies that LMs on different task may share common preferences, making AAR generalize well from single source task to heterogeneous target tasks.

\begin{table*}[t]
\centering
\resizebox{1.0\textwidth}{!}{%
\begin{tabular}{p{4cm}|p{6cm}|p{6cm}}
\hline
   \tf{Question} & \tf{Human-preferred Document} & \tf{LM-preferred Document} \\
    \hline
     what happens if you miss your cruise ship & \emph{\textcolor{red}{If you do miss the ship, go into the cruise terminal and talk with the port agents, who are in contact with both shipboard and shoreside personnel.}} They can help you decide the best way to meet your ... & \emph{\textcolor{green}{The cruise line}} is not financially responsible for getting passengers to the next port if they miss the ship. Your travel to the subsequent port, or home, is on your dime, as are any necessary hotel stays and meals... \\
     \hline
     what is annexation? & \emph{\textcolor{red}{Annexation is an activity in which two things are joined together, usually with a subordinate or lesser thing being attached to a larger thing.}} In strict legal terms, annexation simply involves... & Annexation (Latin ad, to, and nexus, joining) is the administrative action and concept in international law relating to the \emph{\textcolor{green}{forcible transition of one state's territory by another state}}. It is generally held to be an illegal act... \\
    \hline
\end{tabular}
}

\caption{
Cases study on MSMARCO QA dataset. We show Top-1 document annotated by human users and FiDAtt scores. \textcolor{red}{Red} texts are the gold answer spans.
}
\label{tab:cases}
\end{table*}

\subsection{Analysis of LM-preferred Documents}
\label{sec:LM-preferred documents}

\begin{figure}
    \centering
	\begin{subfigure}{0.57\linewidth}
	    \centering
        \includegraphics[width=1.0\linewidth]{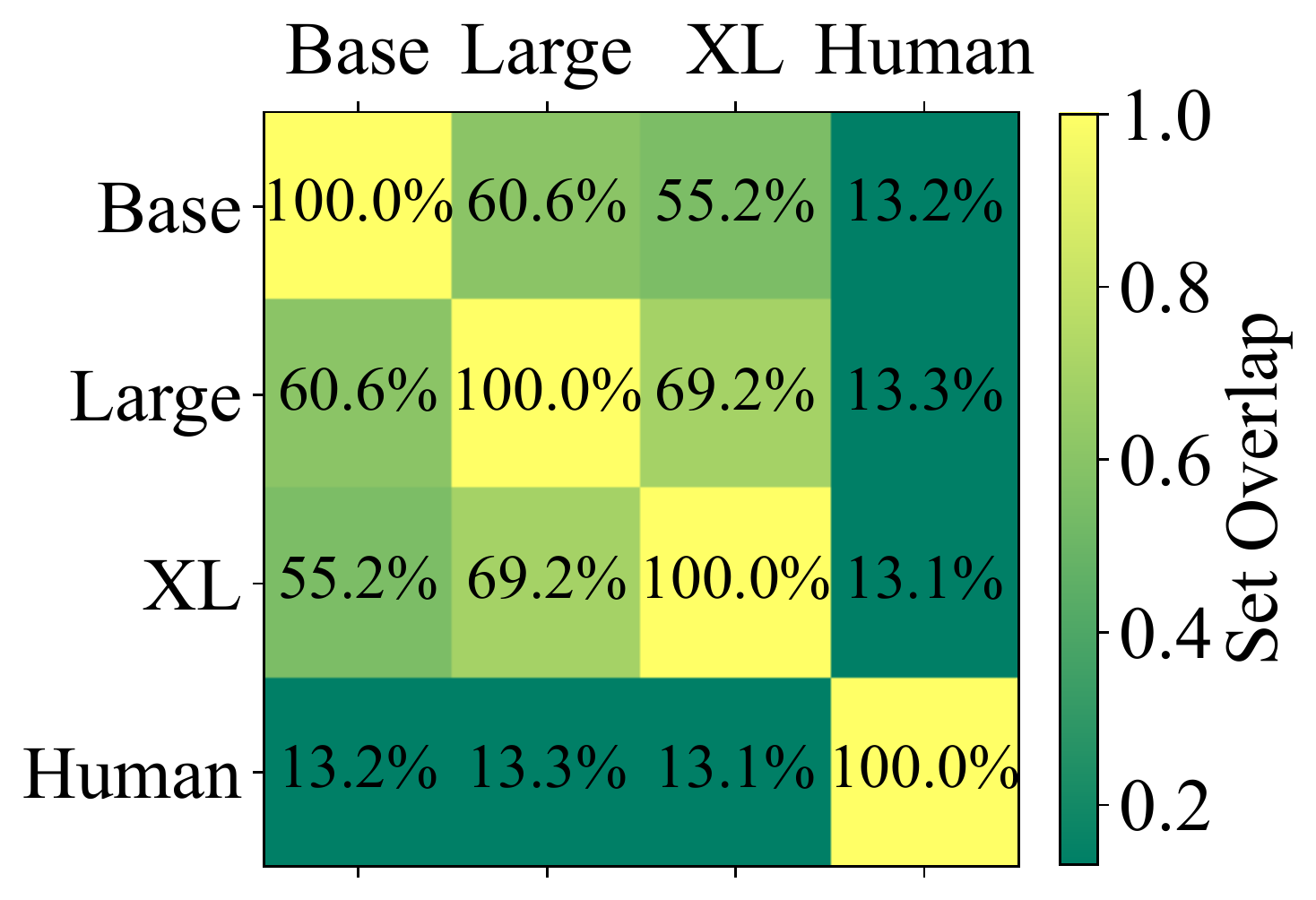}
        \caption{Positive docs overlap.}
        \label{fig-overlap}
	\end{subfigure}%
	\centering
	\begin{subfigure}{0.43\linewidth}
		\centering
        \includegraphics[width=1.0\linewidth]{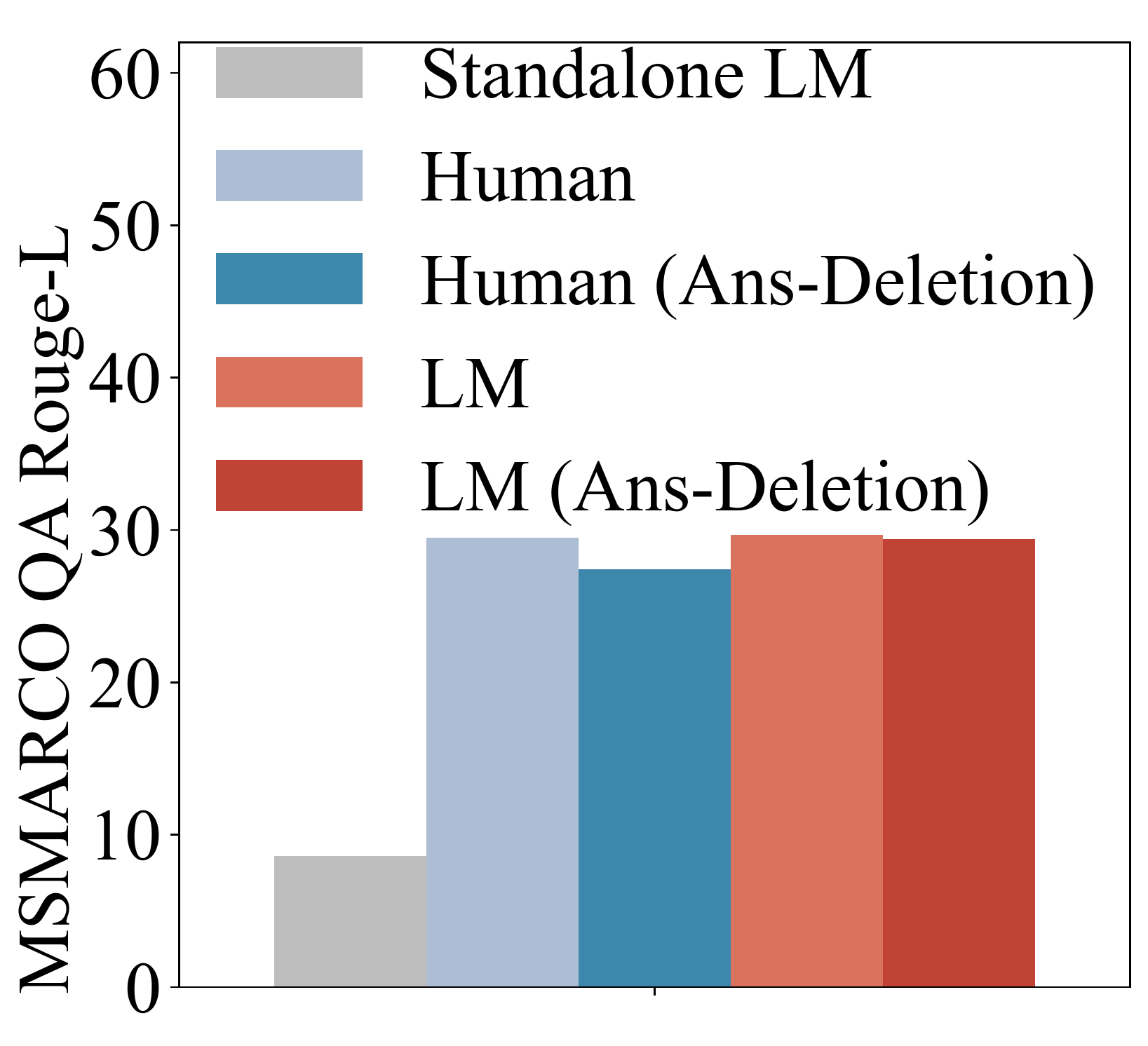}
        \caption{Answer-deletion test.}
        \label{fig-masked}
	\end{subfigure}%
    \caption{Analysis of LM-preferred documents. (a) shows the overlaps of positive document sets, where used LMs are Flan-T5 series. (b) presents the answer-deletion experiments on the MSMARCO QA dataset. The retriever is initialized from ANCE.}
\end{figure}

We highlight the necessity of adapting existing retrievers to LMs by comparing the preferred documents between search users and LMs. In general, we discover that LM-preferred documents can assist LM from alternative perspectives rather than the full information favored by search users.

First, we define the set overlap $O$ between two positive documents set $D^+_{1}$ and $D^+_{2}$ as:
\begin{equation}
\vspace{-1pt}
    O = \frac{D^+_{1} \cap D^+_{2}}{D^+_{1} \cup D^+_{2}}.
\vspace{-1pt}
\end{equation}

As illustrated in Figure~\ref{fig-overlap}, the set overlaps of the positive document sets annotated by human users ($D^{h+}$) and LMs ($\text{Top-}K_{S_i^a,D^a}$) are quite low (near 13\%), demonstrating their distinct tendencies in selecting valuable documents. On the contrary, the overlaps between different LMs are relatively high (over 55\%). This evidence provides a strong rationale for the generalization ability of AAR since LMs with different sizes tend to annotate similar positive documents. Furthermore, LMs whose sizes are closer generally possess higher overlaps. This implies a better generalization ability of the AAR to the LMs whose capacity is near the source LM. The findings further validate the results illustrated in Figure~\ref{fig-positive-selection}.

To give an in-depth analysis of how human-preferred and LM-preferred documents differ, we show two representative cases sampled from the MSMARCO QA in Table~\ref{tab:cases}. We observe that the human-preferred document can always present the gold answer at the beginning of the text, while the LM-preferred document may not contain the exact answer. However, an LM-preferred document can (1) deliver a new perspective to answer the given question, e.g., ``the cruise line's responsibility if you miss your cruise ship'' and (2) give a specific explanation instead of an abstract definition, e.g., ``forcible transition of one state's territory by another state'', These characteristics differ from search users who want the full information and can further assist LMs in knowledge-based reasoning.

\begin{figure}
\centering
\includegraphics[width=0.75\linewidth]
    {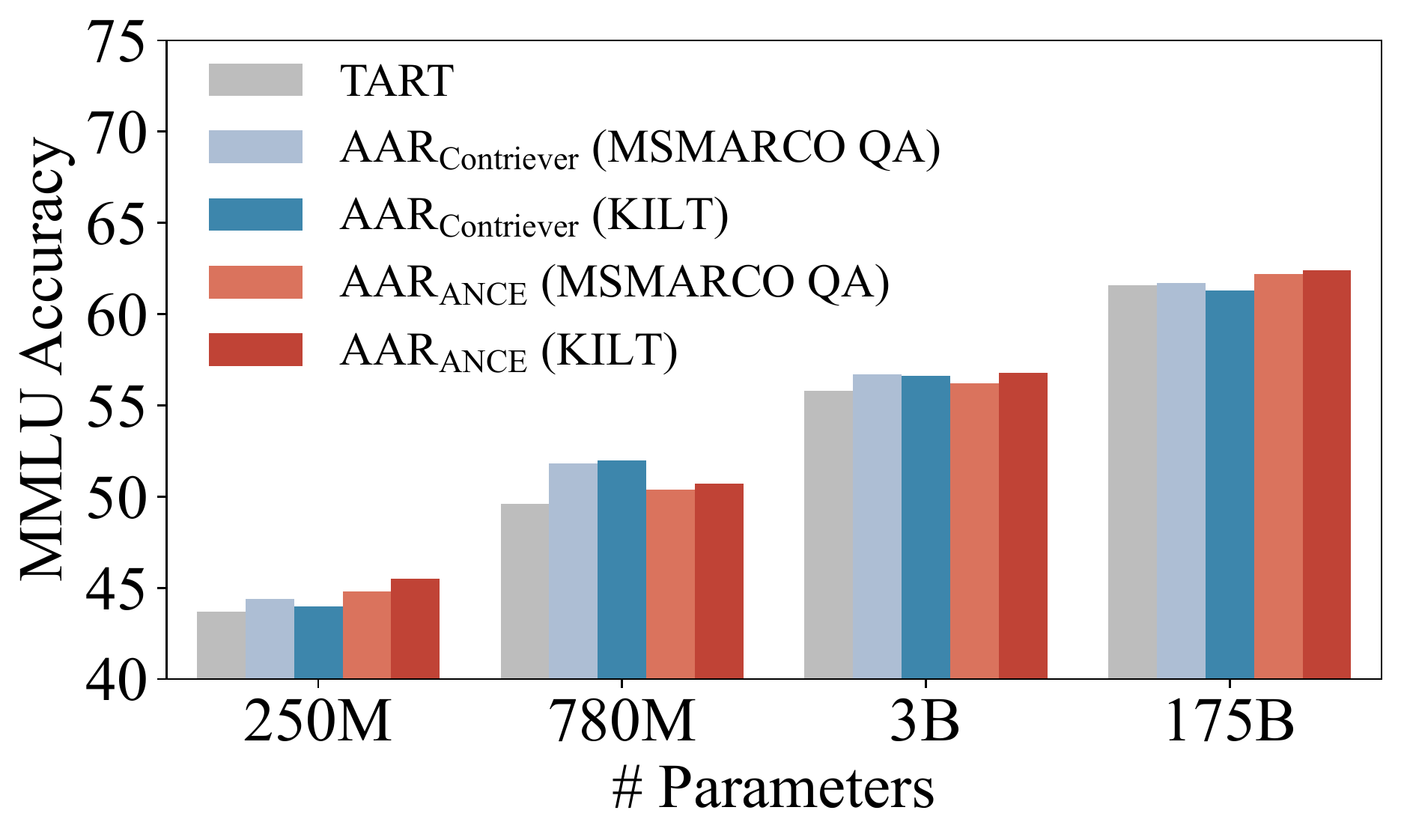}
    \caption{Comparison between single-task (MSMARCO QA) and multi-task (KILT) trained AAR. TART~\cite{asai2022tart} is a multi-task instruction-finetuned retriever that has not been finetuned with LM-preferred signals.}
    \label{fig-multi-task}
\end{figure}

We further examine the unique characteristics of LM-preferred documents through the answer-deletion test (i.e., deleting the exact answer span from the retrieved documents). As shown in Figure~\ref{fig-masked}, the retriever trained by either human-preferred (i.e., human-preferred retriever) or LM-preferred documents (i.e., LM-preferred retriever) can help LM answer the given question. Nevertheless, after the answer-deletion, the performance of LM with the human-preferred retriever declines more significantly than with the LM-preferred retriever. Despite having fewer exact match answers (0.6\% for LM-preferred documents vs. 13.0\% for human-preferred documents), LM-preferred documents provide helpful information from alternative perspectives. Therefore, adapting retrievers with LM-preferred documents can in turn make retrieval-augmented LM perform better.

\subsection{Multi-task Training of AAR}
\label{sec:multi-task training}

In this section, we explore if the multi-task training of AAR can endow the retriever with better generalization to the target task. Specifically, we choose KILT~\cite{fb_kilt} as our multi-task data source, which consists of 5 categories (Fact Checking, Entity Linking, Slot Filling, Open Domain QA, and Dialogue). We take one representative subtask per category to form a mixture of multiple source tasks. 

Figure~\ref{fig-multi-task} illustrates that ANCE trained with multi-task KILT can consistently outperform the single-task MSMARCO QA, proving the better generalization ability brought by multi-task augmentation-adapted training. It is possible that LMs may vary slightly in preferred documents for different tasks and AAR can switch more smoothly to the target task with the help of multi-task training. Contriever does not benefit greatly from multi-task training. We conjecture that this is because Contriever has been pre-trained with multiple formats of data augmentations and thus generalizes better to new data distribution than ANCE. Interestingly, multi-task instruction-finetuned retriever TART~\cite{asai2022tart} has an overall worse performance compared to AAR, highlighting the benefits of having LM-preferred documents during the multi-task training. A more detailed analysis about the selection of source tasks is in Appendix~\ref{sec:select}.

\begin{table}[t]
\centering
\resizebox{1.0\linewidth}{!}{%
\begin{tabular}{l|ccccc|c}
\hline
    \multirow{2}{*}{\tf{Corpora}} & \multicolumn{5}{c|}{\tf{MMLU}} & {\tf{PopQA}} \\ & All & Hum. & Soc. Sci. & STEM & Other & All \\
\hline
    MS MARCO & \tf{44.8} & 42.2 & \tf{46.4} & \tf{39.0} & \tf{53.2} & 13.6 \\
    KILT-Wikipedia & 42.6 & \tf{42.5} & 45.9 & 34.3 & 50.5 & \tf{37.7} \\
    Standalone LM & 36.1 & 40.4 & 39.8 & 27.0 & 40.6 & 8.8 \\
\hline
\end{tabular}
}

\caption{
Ablation of the retrieval corpus, with Flan-T5$_{\text{Base}}$ as LM and AAR$_{\text{ANCE}}$ as retriever.
}
\label{tab:corpus}
\end{table}

\subsection{Effect of Retrieval Corpus}
\label{sec:corpus}

Table~\ref{tab:corpus} demonstrates that regardless of the retrieval corpus, AAR results in consistent and substantial performance gains over the standalone LM.

On MMLU, using MS MARCO as the retrieval corpus improves the LM more compared to KILT-Wikipedia. We hypothesize that the retriever has been trained with MS MARCO corpus and thus holds better retrieval performance on it.

On PopQA, model performance will drop by large margins if we use MS MARCO as the retrieval corpus instead of KILT-Wikipedia. The primary reason is that the PopQA dataset is sampled from Wikidata and designed for long-tail questions. Partial long-tail knowledge can be only found in KILT-Wikipedia~\cite{mallen2023PopQA} while MS MARCO lacks the indispensable evidence that should be utilized for answer prediction. For instance, given the question ``Who is the mother of Melissa Benn?'', there is no document in MS MARCO containing the answer ``Caroline Benn''. Under such circumstances, aligning the retrieval corpus with the data source can be necessary to leverage AAR's ability.

\begin{table}[t]
\centering
\resizebox{1.0\linewidth}{!}{%
\begin{tabular}{ll|c|c}
\hline
\multirow{2}{*}{\tf{Settings}} & \multirow{2}{*}{\tf{Methods}} & \tf{MMLU} & \tf{PopQA} \\
& & All & All \\
\hline
\multirow{2}{*}{Few-shot} & OPT~\cite{zhang2022opt} & 26.0 & 12.3 \\
& GPT-neo~\cite{gpt-neo} & 28.7 & 11.3 \\
\hline
\multirow{8}{*}{Zero-shot} & OPT & 22.7 & 12.0 \\
& GPT-neo & 25.3 & 9.9 \\
& OPT GenRead & 22.3 & 12.2 \\
& GPT-neo GenRead & 24.4 & 11.9 \\
& OPT w/ AAR$_{\text{Contriever}}$ (Ours) & 23.2 & 29.1 \\
& GPT-neo w/ AAR$_{\text{Contriever}}$ (Ours) & 25.2 & 27.8 \\
& OPT w/ AAR$_{\text{ANCE}}$ (Ours) & 23.7 & \tf{32.9} \\
& GPT-neo w/ AAR$_{\text{ANCE}}$ (Ours) & \tf{26.6} & 30.1 \\
\hline
\end{tabular}
}

\caption{
Results of OPT and GPT-neo. We use their 1.3B version. The score marked as \tf{bold} means the best performance in the zero-shot setting.
}
\label{tab:OPT}
\end{table}

\subsection{Application Scenarios of AAR}
\label{sec:scenarios}

To examine if AAR works for unseen LMs that lack zero-shot generalization ability, we also report the results of OPT~\cite{zhang2022opt} and GPT-neo~\cite{gpt-neo}. These models may have poor zero-shot performance due to the lack of multi-task instruction tuning. 

From Table~\ref{tab:OPT}, we find that our AAR improves both LMs marginally on MMLU while achieving significant gains on PopQA. We conjecture that LMs can benefit more easily from retrieval augmentation on the knowledge-probing task like PopQA, where the answer span can be directly acquired from the retrieved documents. MMLU requires the LM to not only comprehend the retrieved pieces of evidence but also perform knowledge-based reasoning over them. OPT and GPT-neo may not possess such abilities in zero-shot scenarios.

In summary, although AAR perfectly fits the multi-task instruction-finetuned LMs such as the Flan-T5 series and InstructGPT, it may not bring significant gains for LMs whose zero-shot performance is sometimes poor, especially on knowledge-based reasoning. However, we believe that multi-task instruction-finetuned models will be the foundation of future work due to their outstanding zero-shot generalization capabilities, ensuring the wide-ranging application scenarios of AAR.
\section{Discussions}


\paragraph{LM-preferred Documents.} Acquiring discrete feedback signals from LMs is challenging as it requires superior labeling ability, which is not the designed purpose of LMs. Inspired by ADist~\cite{izacard2020distilling} and Atlas~\cite{izacard_few-shot_2022}, we utilize the FiDAtt scores to select LM-preferred documents for the augmentation-adapted training. However, FiDAtt scores may not reflect the actual contribution of each document faithfully since LM may prefer attending to readable rather than informative documents. Furthermore, the quality of LM-preferred documents depends heavily on the initial performance of the retrieval-augmented LM. 
Parallel work~\cite{shi2023replug} computes the KL divergence between retrieval likelihood and LM likelihood to train the retriever. Nevertheless, they require a larger source LM, Curie (6.7B), to provide accurate LM likelihood signals. In the future, reinforcement learning could serve as an alternative method to train the retriever, as it optimizes the retriever by directly leveraging LM's signals without relying on the devised rule.

\paragraph{Generic Retrieval Plug-in.}
Chatgpt-retrieval-plugin\footnote{\href{https://github.com/openai/chatgpt-retrieval-plugin}{https://github.com/openai/chatgpt-retrieval-plugin}} has recently gained attention in the NLP community as a generic retrieval plug-in. It retrieves the most relevant document from users' data sources and tailor ChatGPT's response to meet their specific needs. We believe that techniques such as AAR will enhance the ability of black-box ChatGPT to generate more reasonable responses based on the retrieved information, thereby promoting the development of human-centered LM design.
\section{Conclusion and Future Work}


This paper introduces generic retrieval plug-in that utilizes a generic retriever to enhance target LMs that may be unknown in advance or are unable to be fine-tuned jointly.
Our proposed retriever, AAR, can directly support black-box LMs without requiring any fine-tuning of the LMs.
This is accomplished by building the AAR's training data with preferred documents from a small source LM together with the ground truth.

Empirical results on MMLU and PopQA demonstrate that AAR-assisted LMs greatly outperform the standalone ones in zero-shot scenarios, and AAR generalizes well to LMs of different sizes and structures. 
Analytical results reveal that LM-preferred and human-preferred documents complement each other; LM-preferred documents from different LMs overlap significantly, and LMs with similar sizes tend to yield closer document sets.

We leave a more detailed explanation of how different LMs interact with augmentation documents and a more reasonable selection of LM-preferred documents for future work. We hope our work shed light on a path to a generic way of treating large LMs as black boxes and adapting retrievers to augment them.

\section*{Limitations}

Due to the limitation of computational resources, we have not evaluated the Flan-T5$_\text{XXL}$ whose number of parameters is 11B, and the OPT whose number of parameters is greater than 1.3B.

Since OPT and GPT-neo perform poorly in the zero-shot setting and separating attention scores of each document in the input is tedious for decoder-only models, we choose not to use them as source LMs. However, we prove that taking the encoder-decoder model Flan-T5$_\text{Base}$ as our source LM is also robust to augment decoder-only models. We will explore new methods to annotate LM-preferred documents of decoder-only models based on their inherent signals.

\section*{Acknowledgement}

Zichun Yu, Shi Yu, and Zhiyuan Liu are supported by  Institute Guo Qiang at Tsinghua University, Beijing Academy of Artificial Intelligence (BAAI). All authors proposed the original idea together. Zichun Yu conducted the experiments. Zichun Yu, Chenyan Xiong, Shi Yu, and Zhiyuan Liu wrote the paper. Chenyan Xiong and Zhiyuan Liu provided valuable suggestions for the research. We thank Suyu Ge for sharing the ANCE checkpoint initialized from T5$_\text{Base}$.

\clearpage
\bibliography{custom}
\bibliographystyle{acl_natbib}

\clearpage

\appendix

\label{sec:appendix}

\section{Experimental Settings}
\label{sec:setting}

\subsection{Training Hyperparameters}

We take the ANCE initialized from T5$_\text{Base}$\footnote{\href{https://huggingface.co/OpenMatch/t5-ance}{https://huggingface.co/OpenMatch/t5-ance}}~\cite{xiong2021approximate,ge2023augmenting} and Contriever\footnote{\href{https://huggingface.co/facebook/contriever-msmarco}{https://huggingface.co/facebook/contriever-msmarco}}~\cite{izacard2021contriever}'s hyperparameters in the augmentation-adapted training. Specifically, we fix batch size as 8, learning rate as 5e-6, and epochs as 6 for ANCE while taking batch size as 8, learning rate as 1e-5, and epochs as 3 for Contriever. We choose their best checkpoints based on the performance of the development set. The information about our source tasks and target tasks are listed in Table~\ref{tab:task-info}.

\subsection{Number of Augmentation Documents}

LMs of different sizes, facing various target tasks, may require indefinite numbers of augmentation documents to achieve their best performance.

\begin{figure*}
	\centering
	\begin{subfigure}{0.28\textwidth}
	    \centering
        \includegraphics[width=1.0\linewidth]{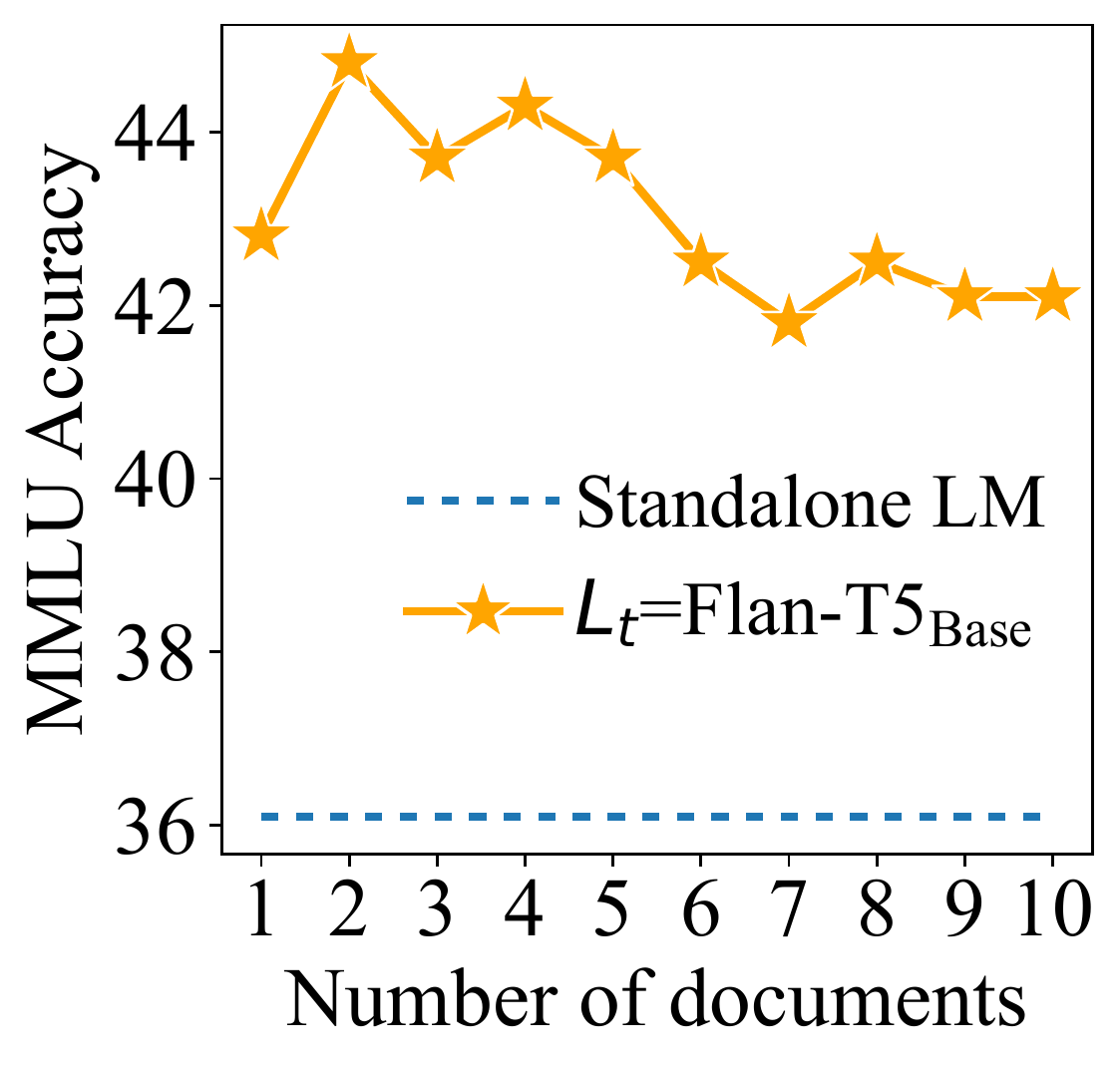}
        \caption{Flan-T5$_{\text{Base}}$ w/ AAR$_{\text{ANCE}}$.}
	\end{subfigure}%
	\centering
	\begin{subfigure}{0.28\textwidth}
		\centering
        \includegraphics[width=1.0\linewidth]{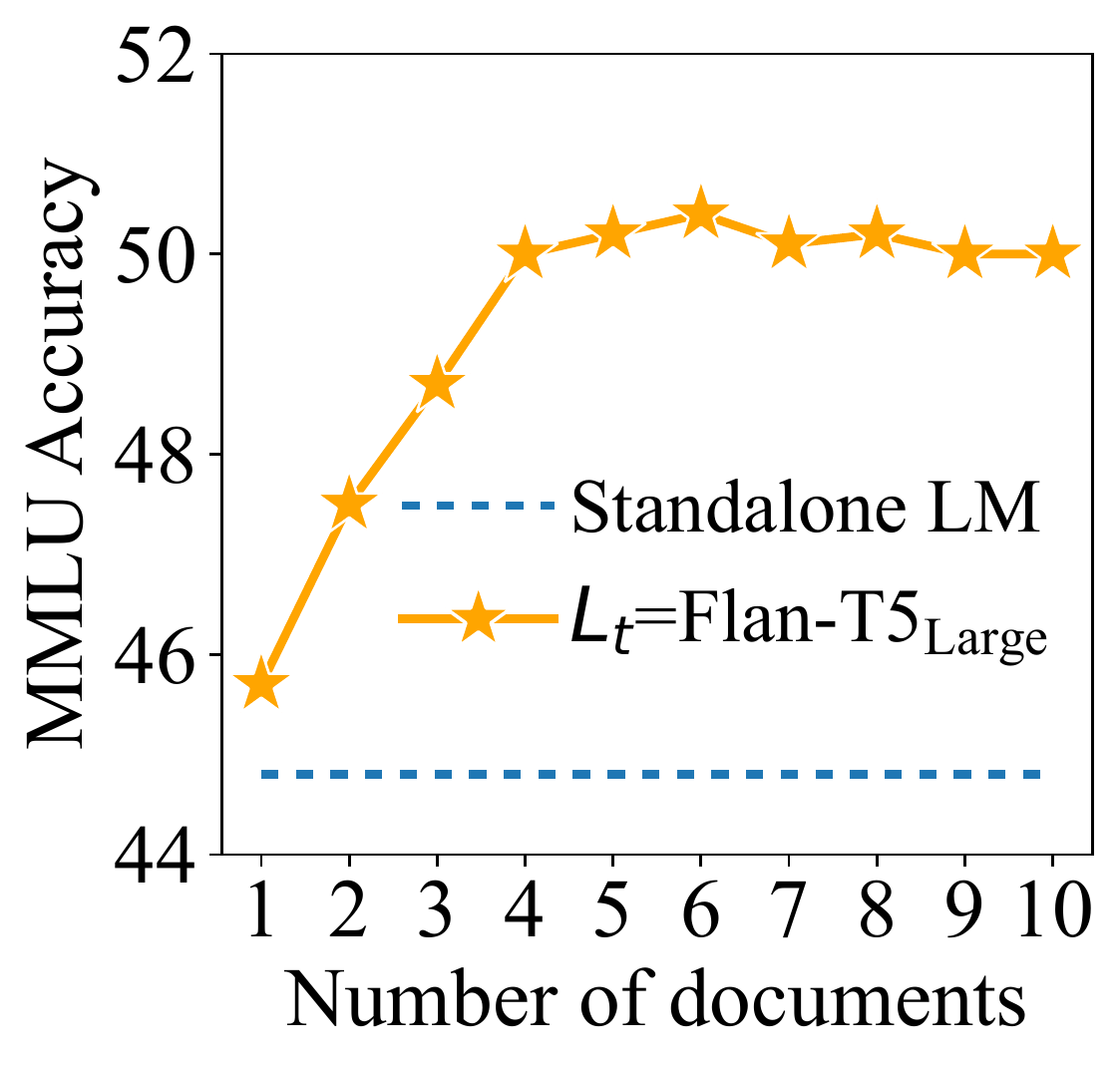}
        \caption{Flan-T5$_{\text{Large}}$ w/ AAR$_{\text{ANCE}}$.}
	\end{subfigure}%
	\centering
	\begin{subfigure}{0.28\textwidth}
	    \centering
        \includegraphics[width=1.0\linewidth]{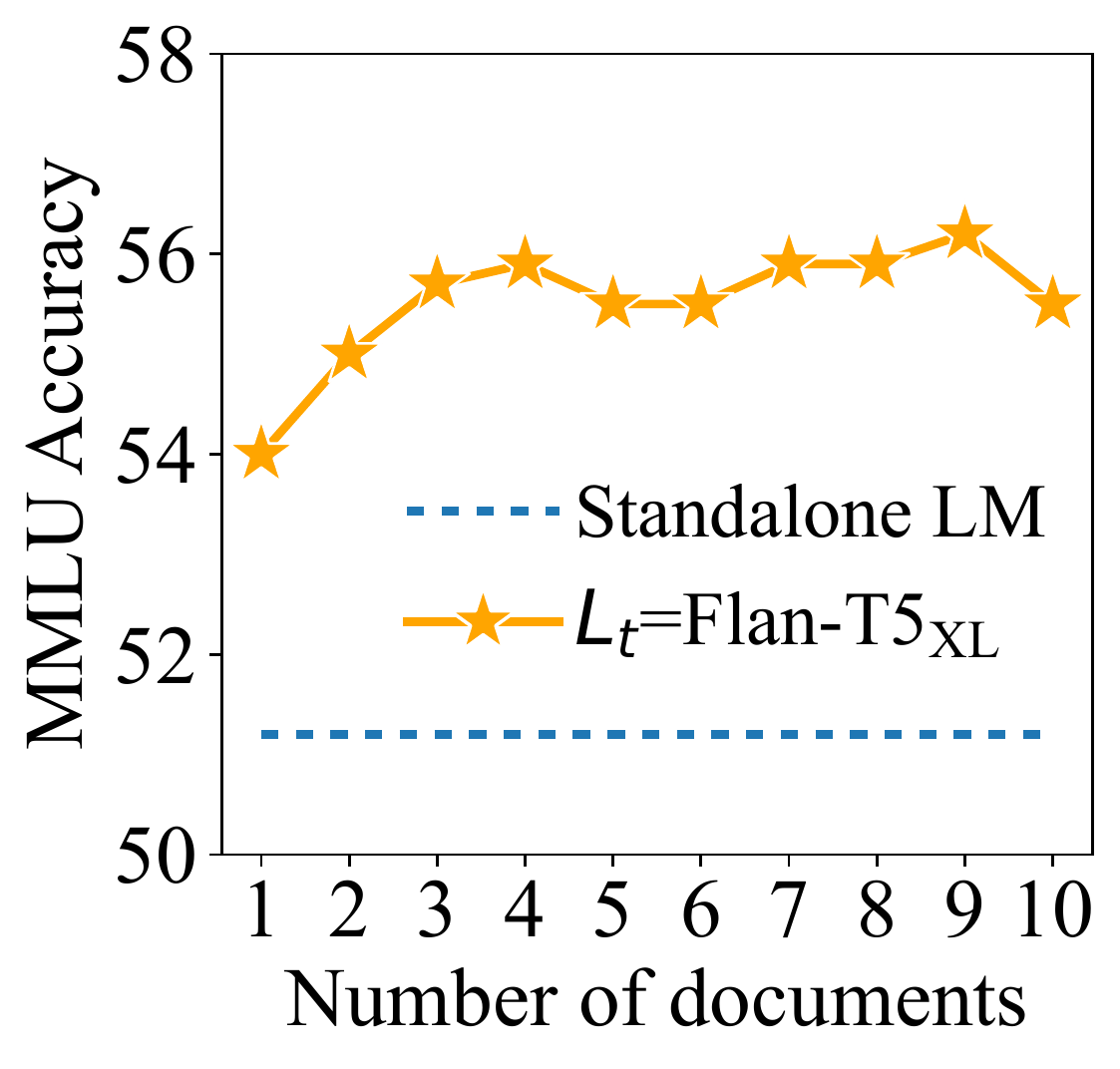}
        \caption{Flan-T5$_{\text{XL}}$ w/ AAR$_{\text{ANCE}}$.}
	\end{subfigure}%
	\centering
	\caption{Relationship between LM's performance and the number of augmentation documents.}
	\label{fig-relation}
\end{figure*}

For MMLU, we analyze how the number of augmentation documents affects LMs' performance. As illustrated in Figure~\ref{fig-relation}, we discover that LMs of larger capacity generally benefit more from more augmentation documents. A possible explanation is that larger LMs are more capable of integrating information from multiple documents and performing complicated reasoning based on them.

For PopQA, using 3 augmentation documents achieves the best performance across all LMs.


\subsection{Prompt Templates}

The prompt template for MMLU is:

\texttt{Here's a problem to solve: \{question\}}

\texttt{Among the 4 following options, which is the correct answer?}

\texttt{- A: \{choice\_A\}}

\texttt{- B: \{choice\_B\}}

\texttt{- C: \{choice\_C\}}

\texttt{- D: \{choice\_D\}}

The prompt template for PopQA is:

\texttt{Q: \{question\} A:}

\begin{table}[ht]
\centering
\resizebox{0.77\linewidth}{!}{%
\begin{tabular}{l|c|c|c}
\hline
    \diagbox[width=85pt,height=20pt]{$T_s$}{$T_t$} & MMLU & NQ & zsRE \\
\hline
    MSMARCO QA & \tf{44.8} & \tf{46.7} & 75.1  \\
    KILT-TriviaQA & 43.6 & 46.4 & 74.9 \\
    KILT-T-REx & 44.1 & 45.9 & \tf{77.2} \\
\hline
\end{tabular}
}

\caption{
    Relationship between the selection of source task $T_s$ and the performance of target task $T_t$. The model is Flan-T5$_{\text{Base}}$ w/ AAR$_{\text{ANCE}}$. As NQ and zsRE are included in the Flan-T5 training data, we only report their F1 results here for reference.
}
\label{tab:st}
\end{table}

\section{Selection of Source Task}
\label{sec:select}

We provide a detailed selection of the source tasks here, using a variety of source and target tasks to analyze. MSMARCO QA, KILT-TriviaQA, and NQ belong to Open Domain QA, while KILT-T-REx and zsRE belong to Slot Filling. MMLU belongs to Multi-task Language Understanding, which is closer to the Open Domain QA in terms of the task objective. As shown in Table~\ref{tab:st}, when we align the category of the source task with the target task, the LM w/ AAR can generally achieve the best results. We suppose that this is because LM may share similar document preferences on the tasks from the same dataset category, making AAR easier to generalize. Furthermore, taking MSMARCO QA as the source task performs the best on MMLU. This validates the rationality to set $T_s$ as MSMARCO QA in our main experimental settings.

\section{AAR's Improvements on PopQA}

\begin{figure}[h]
\centering
\includegraphics[width=0.8\linewidth]
    {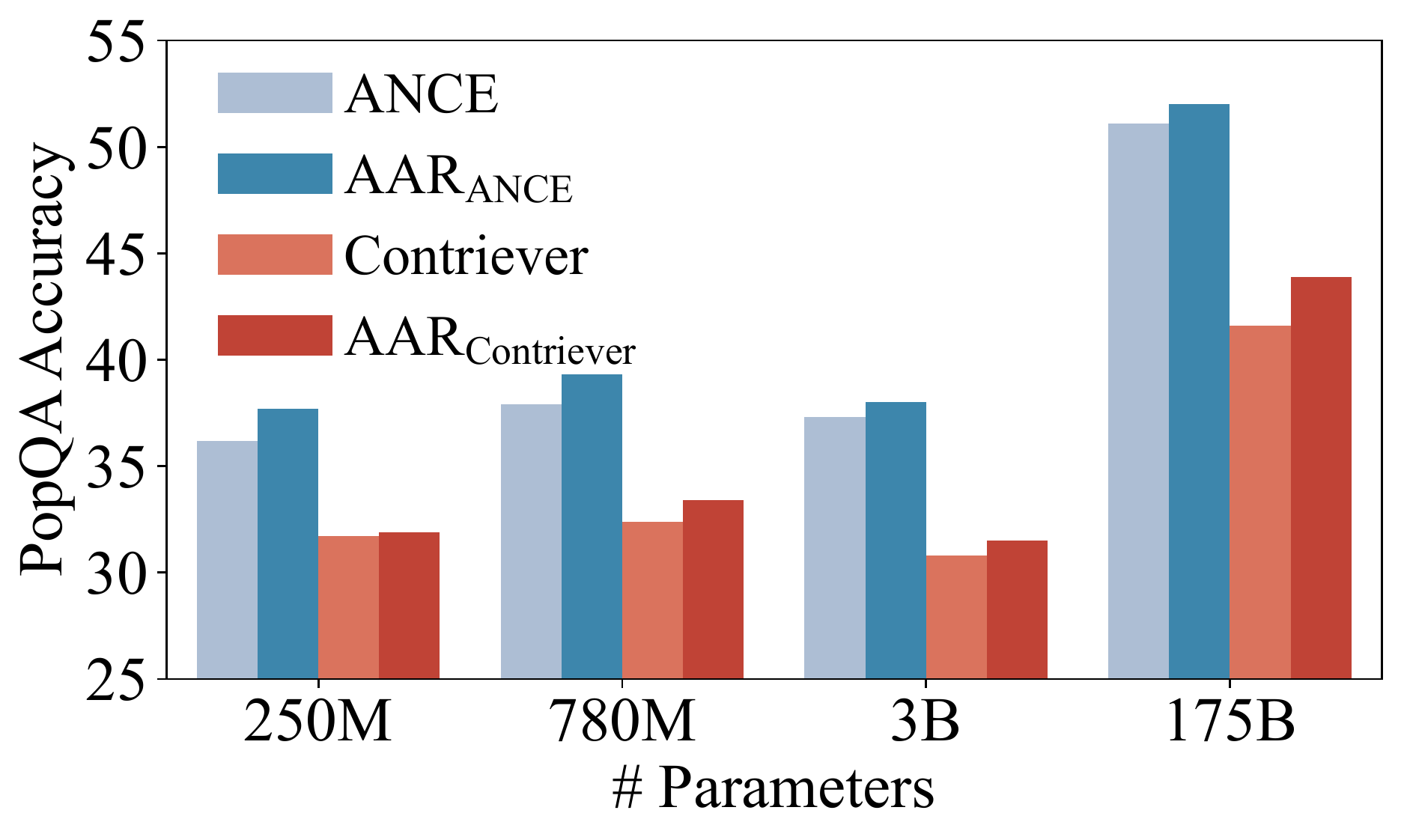}
    \caption{AAR's improvements on PopQA, using Flan-T5$_{\text{Base}}$ (250M), Flan-T5$_{\text{Large}}$ (780M), Flan-T5$_{\text{XL}}$ (3B), InstructGPT (175B) as target LMs.}
    \label{fig-popqa-improvement}
\end{figure}

\begin{table*}
\centering
\resizebox{0.8\textwidth}{!}{%
\begin{tabular}{ll|c|c}
\hline
& & \tf{Category} & \tf{Number}  \\
\hline
\multirow{5}{*}{$T_s$} & MSMARCO QA & Open Domain QA & 148122  \\
& KILT-FEVER & Fact Checking & 10444 \\
& KILT-WNED & Entity Linking & 3396 \\
& KILT-T-REx & Slot Filling	& 5000 \\
& KILT-TriviaQA & Open Domain QA & 5359 \\
& KILT-Wizard of Wikipedia & Dialogue & 3054 \\
\hline
\multirow{2}{*}{$T_t$} 
& MMLU & Multi-task Language Understanding & 1531 \\
& PopQA & Open Domain QA & 14267 \\
\hline
\end{tabular}
}

\caption{
Configurations of our source tasks and target tasks.
}
\label{tab:task-info}
\end{table*}

\begin{table*}
\centering
\resizebox{0.8\textwidth}{!}{%
\begin{tabular}{l|ccccc}
\hline 
\multirow{2}{*}{\tf{Methods}} & \multicolumn{5}{c}{\tf{MMLU}} \\ & All & Hum. & Soc. Sci. & STEM & Other \\
\hline
Flan-T5$_{\text{Base}}$ & 36.1 & 40.4 & 39.8 & 27.0 & 40.6 \\
Flan-T5$_{\text{Base}}$ Fine-tuning & 36.1 & 38.9 & 41.2 & 27.9 & 39.9 \\
Flan-T5$_{\text{Base}}$ w/ Contriever & 43.7 & 44.4 & 45.0 & 36.4 & 51.1 \\
Flan-T5$_{\text{Base}}$ w/ ANCE & 43.0 & 44.2 & 44.3 & 34.5 & 51.9 \\
Flan-T5$_{\text{Base}}$ w/ AAR$_{\text{Contriever}}$ (Ours) & 44.4 & \tf{44.7} & \tf{47.7} & 35.8 & 52.2 \\
Flan-T5$_{\text{Base}}$ w/ AAR$_{\text{ANCE}}$ (Ours) & \tf{44.8} & 42.2 & 46.4 & \tf{39.0} & \tf{53.2} \\
\hline
Flan-T5$_{\text{Large}}$ & 45.1 & 47.7 & 53.5 & 34.4 & 49.2 \\
Flan-T5$_{\text{Large}}$ Fine-tuning & 45.3 & 47.6 & 54.1 & 35.2 & 48.7 \\
Flan-T5$_{\text{Large}}$ w/ Contriever & 50.7 & 50.5 & 56.4 & 38.9 & 61.1 \\
Flan-T5$_{\text{Large}}$ w/ ANCE & 49.2 & 49.3 & 56.7 & 38.1 & 57.2 \\
Flan-T5$_{\text{Large}}$ w/ AAR$_{\text{Contriever}}$ (Ours) & \tf{51.8} & \tf{50.8} & \tf{59.7} & \tf{39.4} & \tf{61.8} \\
Flan-T5$_{\text{Large}}$ w/ AAR$_{\text{ANCE}}$ (Ours) & 50.4 & 48.0 & 58.1 & 39.3 & 60.2 \\
\hline
Flan-T5$_{\text{XL}}$ & 51.2 & 55.5 & 57.4 & 38.1 & 58.7 \\
Flan-T5$_{\text{XL}}$ w/ Contriever & 56.4 & 57.3 & \tf{66.1} & \tf{43.9} & 63.2 \\
Flan-T5$_{\text{XL}}$ w/ ANCE & 55.3 & 55.9 & 64.0 & 41.5 & 64.9 \\
Flan-T5$_{\text{XL}}$ w/ AAR$_{\text{Contriever}}$ (Ours) & \tf{56.7} & 57.7 & 65.4 & 43.6 & \tf{65.1} \\
Flan-T5$_{\text{XL}}$ w/ AAR$_{\text{ANCE}}$ (Ours) & 56.2 & \tf{59.4} & 64.8 & 41.5 & 64.9 \\
\hline
InstructGPT & 60.2 & \tf{65.7} & 68.0 & 46.1 & 66.5 \\
InstructGPT w/ Contriever & 60.5 & 62.0 & 71.8 & 44.3 & 70.1 \\
InstructGPT w/ ANCE & 61.6 & 62.4 & \tf{73.4} & 47.6 & 68.6 \\
InstructGPT w/ AAR$_{\text{Contriever}}$ (Ours) & 61.5 & 64.5 & 73.1 & 45.0 & 69.9 \\
InstructGPT w/ AAR$_{\text{ANCE}}$ (Ours) & \tf{62.2} & 62.0 & 72.0 & \tf{49.2} & \tf{70.7} \\
\hline
\end{tabular}
}

\caption{
Fine-tuning results on MMLU. We use the official auxiliary training data of MMLU to fine-tune the LM.
}
\label{tab:ft}
\end{table*}

\section{Fine-tuning Results}
We also report the fine-tuning results of Flan-T5$_{\text{Base}}$ and Flan-T5$_{\text{Large}}$ on MMLU auxiliary training data~\cite{hendryckstest2021} in Table~\ref{tab:ft}. Due to the limitation of the computational resources, we do not include the fine-tuning result of Flan-T5$_{\text{XL}}$. We take batch size as 32, learning rate as 5e-5, and epochs as 3 in fine-tuning. In general, the LM that has already been massively multi-task instruction-finetuned, such as Flan-T5, improves little from fine-tuning on extra tasks but benefits greatly from our AAR. 
The results further validate the power of zero-shot retrieval augmentation.

\end{document}